\documentclass[english]{IEEEtran}
\usepackage[T1]{fontenc}
\usepackage[latin9]{inputenc}
\usepackage{babel}
\usepackage{array}
\usepackage{rotating}
\usepackage{url}
\usepackage{multirow}
\usepackage{amsmath}
\usepackage{graphicx}
\usepackage[authoryear]{natbib}
\usepackage[unicode=true]
 {hyperref}

\makeatletter

\providecommand{\tabularnewline}{\\}

 \usepackage{multicol}
\date{}
\usepackage{caption}

\makeatother

\begin{document}

\title{LABR: A Large Scale Arabic Sentiment Analysis Benchmark}

\author{\center
\begin{multicols}{3} 
\center \textbf{Mahmoud Nabil}\\Dept of Computer Engineering \\Cairo University\\Giza, Egypt\\mah.nabil@ieee.org\\ 
\center \textbf{Mohamed Aly}\\Dept of Computer Engineering \\Cairo University\\Giza, Egypt\\mohamed@mohamedaly.info\\
\center \textbf{Amir F. Atiya}\\Dept of Computer Engineering \\Cairo University\\Giza, Egypt\\amir@alumni.caltech.edu\\

\end{multicols}}
\maketitle
\begin{abstract}
We introduce LABR, the largest sentiment analysis dataset to-date
for the Arabic language. It consists of over 63,000 book reviews,
each rated on a scale of 1 to 5 stars. We investigate the properties
of the dataset, and present its statistics. We explore using the dataset
for two tasks: (1) sentiment polarity classification; and (2) ratings
classification. Moreover, we provide standard splits of the dataset
into training, validation and testing, for both polarity and ratings
classification, in both balanced and unbalanced settings. We extend
our previous work by performing a comprehensive analysis on the dataset.
In particular, we perform an extended survey of the different classifiers
typically used for the sentiment polarity classification problem.
We also construct a sentiment lexicon from the dataset that contains
both single and compound sentiment words and we explore its effectiveness.
We make the dataset and experimental details publicly available.
\end{abstract}

\section{Introduction}

The internet is full of platforms where users can express their opinions
about different subjects, from movies and commercial products to books
and restaurants. With the explosion of social media, this has become
easier and more prevalent than ever. Mining these troves of unstructured
text has become a very active area of research with lots of applications.
Sentiment classification is among the most studied tasks for processing
opinions \citet{pang2008opinion}. In its basic form, it involves
classifying a piece of opinion, e.g. a movie or book review, into
either having a positive or negative sentiment. Another form involves
predicting the actual rating of a review, e.g. predicting the number
of stars on a scale from 1 to 5 stars. On the other hand, the goal
of feature-based opinion mining is to identify the main entity in
the review (e.g. a review of a movie or a product) or analyze the
attitude towards a certain aspect of the review (e.g. the performance
of one actor or the battery life of a camera). 

A lot of work has been proposed that target most of the challenging
aspects of the sentiment analysis task, \citep{maynard2012challenges}
discuss some of these challenges. These challenges have some similarities
across different languages but there are many issues and problems
that are specific to each language.

Most of the work done on sentiment analysis and the data sets gathered
target the English language with very little work on Arabic. One of
the reasons is the prevalence of English websites where 55\% of the
visited websites on the Internet use English\footnote{\href{http://en.wikipedia.org/wiki/Languages_used_on_the_Internet}{Languages used on the Internet}

http://en.wikipedia.org/wiki/Languages\_used\_on\_the\_Internet}. Another reason is the complexities of the Arabic language and the
different Arabic dialects existing in each Arab country and even within
each Arab country. However, Arabic is the sixth most widely spoken
language, and therefore it is important to develop sentiment analysis
tools for it. In this work, we set out to address the lack of large-scale
Arabic sentiment analysis datasets in this field, in the hope of sparking
more interest in research in Arabic sentiment analysis and related
tasks. 

Towards this end, we introduce \textbf{LABR}, the \textbf{L}arge-scale
\textbf{A}rabic \textbf{B}ook \textbf{R}eview dataset. It is a set
of over 63,000 book reviews, each with a rating of 1 to 5 stars. This
paper provides a more comprehensive analysis with further contributions
than our preliminary work \citet{alyatyia}. In particular, we experiment
with an expanded set of classifiers to establish a baseline benchmark
that future algorithms can compare to. We also describe a new approach
of extracting a domain specific sentiment lexicon from the training
set, that can help in reducing the time and space complexity of classification.
The dataset and scripts to reproduce the experiments mentioned in
this work can be found online at \url{http://www.mohamedaly.info/datasets/labr}.
\begin{quotation}
The contributions in this paper can be summarized as follows:\end{quotation}
\begin{enumerate}
\item We present the largest Arabic sentiment analysis dataset to-date (up
to our knowledge). 
\item We provide standard splits for the dataset into training, validation
and testing sets. This will make comparing different results much
easier. All the splits and scripts to reproduce the experiments in
this paper are publicly available online.
\item We apply a wide range of classifiers to the large set of book reviews
that we collected. This provides a standard baseline that can be used
by future approaches for comparison purposes.
\item We construct a seed sentiment lexicon from the dataset, and explore
its properties and effectiveness.
\end{enumerate}

\section{Related Work}

Sentiment analysis is handled by either lexicon-based approaches,
machine learning approaches like text classification tasks, or hybrid
approaches \citep{pang2008opinion}.For lexicon-based approaches,
\citep{taboada2011lexicon} developed a Semantic Orientation CALculator
and used some annotated dictionaries of words where the annotation
covers the word polarity and strength. They used Amazon\textquoteright s
Mechanical Turk service to collect validation data to their dictionaries
and based their experiments on four different corpora with equal numbers
of positive and negative reviews. \citep{gindl2010cross} and \citep{ding2008holistic}
used a sentiment lexicon that depends on the context of every polarity
word (contextualized sentiment lexicon) and based their experiments
on customer reviews from Amazon and TripAdvisor\footnote{\href{http://www.tripadvisor.com}{Trip Advisor}}.In
general lexicon-based sentiment classifiers show a positive bias \citep{kennedy2006sentiment},
however \citep{voll2007not} implemented normalization techniques
to overcome this bias.

For machine learning approaches, \citep{pak2010twitter} used part
of speech and n-grams to build sentiment classifiers using the multinomial
naive Bayes classifier, SVM and conditional random fields. They tested
their classifiers on a set of hand annotated twitter posts. \citep{Jiang2011}
proposed an approach to target dependent features in the review by
incorporating syntactic features that are related to the sentiment
target of the review. They built a binary SVM classifier to perform
the classification of two tasks: subjectivity classification and polarity
classification.

For hybrid approaches, \citep{kouloumpis2011twitter} used n-gram
features, lexicon features, and part of speech to build an Ada-boost
classifier. They used three different corpora of Twitter messages
(HASH, EMOT and iSieve) to evaluate their system. \citep{grabner2012classification}constructed
a domain specific lexicon and used it to back the classification of
the reviews. They used a data set for customer reviews from TripAdvisor. 

Concerning the Arabic language, little work has considered the sentiment
analysis problem. \citep{abbasi2008sentiment} performed a multilingual
sentiment analysis of English and Arabic Web forums. \citep{abdul2014samar}
proposed the SAMAR system that perform subjectivity and sentiment
analysis for Arabic social media using some Arabic morphological features.
\citet{abdul2012toward} proposed a way to expand a modern standard
Arabic polarity lexicon from an English polarity lexicon using a simple
machine translation scheme. \citet{elhawary2010mining} built a system
that mines Arabic business reviews obtained from the internet. Also,
they built a sentiment lexicon using a seed list of sentiment words
and an Arabic similarity graph. \citet{shoukry2012preprocessing}
tested the effect of some Arabic preprocessing steps (normalization,
stemming, and stop words removal) on the performance of an Arabic
sentiment analysis system. Simultaneous to our work on sentiment lexicon
generation \citet{elsahar2015building} proposed a method based on
the SVM classifier. Their system, which has some similarities with
our lexicon generation approach has been independently developed. 

Some Arabic sentiment data sets have been collected as follows (summarized
in Table \ref{tab:Arabic-sentiment-data}):

\textbf{OCA }Opinion Corpus for Arabic \citet{rushdi2011oca} contains
500 movie reviews in Arabic, collected from forums and websites. It
is divided into 250 positive and 250 negative reviews, although the
division is not standard in that there is no rating for neutral reviews.
It provides a 10-star rating system, where ratings above and including
5 are considered positive and those below 5 are considered negative.

\textbf{AWATIF} is a multi-genre corpus for Modern Standard Arabic
sentiment analysis \citet{abdul2012awatif}, It contains 2855 reviews
collected from wikipedia talk pages and forums.

\textbf{TAGREED (TGRD), TAHRIR (THR) and MONTADA (MONT) }\citet{abdul2014samar}
used the three corpora to evaluate SAMAR system (A System for Subjectivity
and Sentiment Analysis).

These datasets, however, have a few problems. First, they are considerable
small, with the largest having over 3,000 examples. Second, most of
them are not publicly available. Third, they do not have standard
splits into training and testing, that can provide a standard benchmark
for future research. LABR covers all these weaknesses and provides
a dataset that is an order of magnitude larger and publicly available
with standard benchmarks and baseline experiments.

\begin{table*}[t]
\center
\scalebox{0.85}{%
\begin{tabular}{|c|c|c|c|c|}
\hline 
\textbf{Data Set Name} & \textbf{Size} & \textbf{Source} & \textbf{Type} & \textbf{Cite}\tabularnewline
\hline 
\hline 
TAGREED (TGRD) & 3015 & Tweets & MSA/Dialectal & \citet{abdul2014samar}\tabularnewline
\hline 
TAHRIR (THR) & 3008 & Wikipedia TalkPages  & MSA & \citet{abdul2014samar}\tabularnewline
\hline 
MONTADA (MONT) & 3097 & Forums & MSA/Dialectal & \citet{abdul2014samar}\tabularnewline
\hline 
OCA (Opinion Corpus for Arabic) & 500 & Movie reviews & Dialectal & \citet{rushdi2011oca}\tabularnewline
\hline 
AWATIF & 2855 & Wikipedia TalkPages/Forums & MSA/Dialectal & \citet{abdul2012awatif}\tabularnewline
\hline 
LABR(Large Scale Arabic Book Reviews) & 63,257 & GoodReads reviews\footnote{www.goodreads.com} & MSA/Dialectal & \citet{alyatyia}\tabularnewline
\hline 
\end{tabular}

}

\protect\caption{\textbf{Arabic Sentiment Datasets}.\label{tab:Arabic-sentiment-data}}
\end{table*}

\section{Sentiment Analysis Challenges }

Sentiment analysis is still a formidable natural language processing
task \citet{maynard2012challenges} because unlike text categorization
where the tokens depend largely on the domain or the category, in
sentiment analysis we usually have three semantic orientations (positive,
negative, and neutral) and most tokens can exist in the three categories
at the same time. Another reason is the language ambiguity where one
or more polarity token depends on the context of the sentence. Also
many Internet users tend to give a positive rating even if their reviews
contain some misgivings about the entity, or some sort of sarcastic
remarks, where the intent of the user is the opposite of the written
text.

Some challenges are specific to Arabic language such as few research
work \citet{abbasi2008sentiment}; \citet{Abdul-Mageed2011}; \citet{Abdul-Mageed2011a};
\citet{abdul2012awatif}, and very few available datasets for different
natural language processing tasks. In addition, the complexities of
the Arabic language, due to Arabic being a morphologically rich language,
add a level of complication (see \citet{el2009kp} and \citet{el2011accuracy}).
Another problem is the existence of Modern Standard Arabic side by
side with different Arabic dialects, which are not yet standardized.
\citet{el2013open} discussed some other challenges specific to the
Arabic language such as the unavailability of colloquial Arabic parsers.
This is a problem that plagues all work that depend on the parsed
structure of the sentence. Also there is a need for person named entity
recognition as some Arabic names are derived from adjectives. Another
problem is that the sentiment of compound phrases is often not related
to that of its constituent words. To deal with all these challenges
this work propose the largest Arabic sentiment analysis dataset with
standard splits in order to make comparing different results easier.
Also we provide a set of baseline sentiment analysis experiments to
the data set and finally, we propose a method to construct a seed
sentiment lexicon from the dataset.

\section{Dataset Collection and Properties\label{sec:Dataset-Collection-and-Properties}}

\begin{figure*}[t]
\center\includegraphics[scale=0.68]{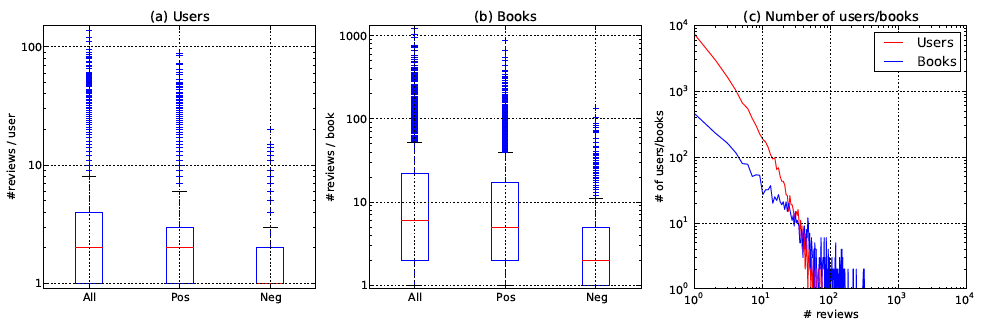}

\protect\caption{\textbf{Users and Books Statistics}.\label{fig:Number-of-reviews-per-user}
\emph{(a) }Box plot of the number of reviews per user for all, positive,
and negative reviews. The \emph{red} line denotes the median, and
the edges of the box the \emph{quartiles}. \emph{(b)} the number of
reviews per book for all, positive, and negative reviews. \emph{(c)}
the number of books/users with a given number of reviews.}
\end{figure*}

\begin{figure*}[t]
\center\includegraphics[scale=0.69]{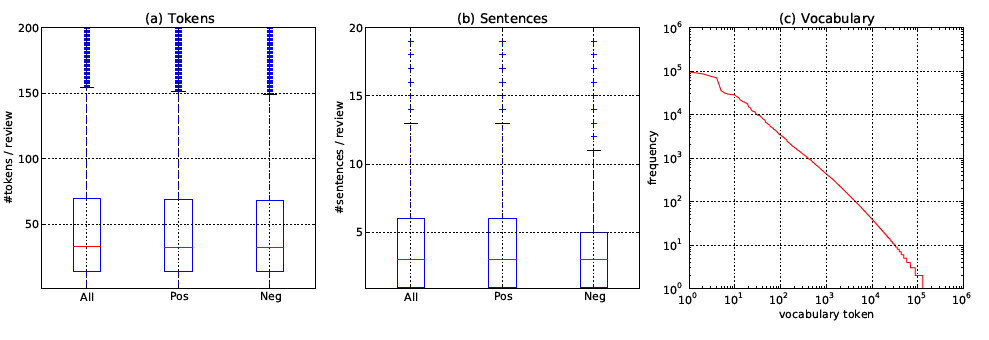}

\protect\caption{\textbf{Tokens and Sentences Statistics} \label{fig:Tokens-and-Sentences}.
\emph{(a)} the number of tokens per review for all, positive, and
negative reviews. \emph{(b)} the number of sentences per review. \emph{(c)}
the frequency distribution of the vocabulary tokens.}
\end{figure*}

\begin{table}
\center
\scalebox{0.95}{%
\begin{tabular}{|c|c|}
\hline 
Number of reviews & 63,257\tabularnewline
\hline 
\hline 
Number of users & 16,486\tabularnewline
\hline 
Avg. reviews per user & 3.84\tabularnewline
\hline 
Median reviews per user & 2\tabularnewline
\hline 
Number of books & 2,131\tabularnewline
\hline 
Avg. reviews per book & 29.68\tabularnewline
\hline 
Median reviews per book & 6\tabularnewline
\hline 
Median tokens per review & 33\tabularnewline
\hline 
Max tokens per review & 3,736\tabularnewline
\hline 
Avg. tokens per review & 65\tabularnewline
\hline 
Number of tokens & 4,134,853\tabularnewline
\hline 
Number of sentences & 342,199\tabularnewline
\hline 
\end{tabular}

}

\protect\caption{\textbf{Important Dataset Statistics}.\label{tab:Important-Dataset-Statistics.}
See section \ref{sec:Dataset-Properties}.}
\end{table}

\begin{figure}
\center\includegraphics[scale=0.69]{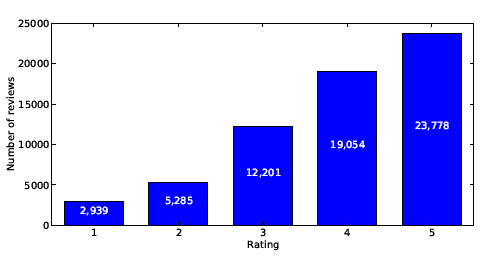}

\protect\caption{\label{fig:Reviews-Histogram}\textbf{Reviews Histogram}: The number
of reviews for each rating. Notice the unbalance in the dataset, with
much more positive reviews (ratings 4 and 5) than negative (ratings
1 and 2) or neutral (rating 3). See section \ref{sec:Dataset-Properties}.}
\end{figure}

\begin{figure}
\center
\scalebox{0.99}{\includegraphics[width=0.43\paperwidth,height=0.3\paperheight]{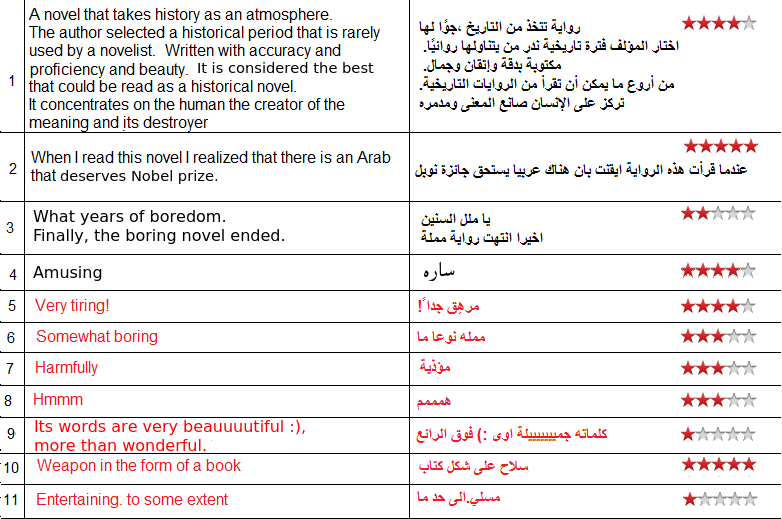}

}

\protect\caption{\textbf{LABR reviews examples}.\label{fig:LABR-reviews-examples}
The English translation is in the left column, the original Arabic
review on the right, and the rating shown in stars. Notice the noise
in some of the ratings, for example reviews 4, 9, and 11. Notice also
the ambiguity for the reviews with rating 3, which can be associated
with positive, negative, or neutral. See also section \ref{sec:Dataset-Properties}.}
\end{figure}

\subsection{Dataset Collection\label{sec:Dataset-Collection}}

We downloaded over 220,000 reviews from the book readers social network
\url{ www.goodreads.com} during the month of March 2013. These reviews
were from the first 2,143 books in the list of Best Arabic Books.
After harvesting the reviews, we found out that over 70\% of them
were not in Arabic, either because some non-Arabic books or translations
of Arabic books to other languages exist in the list. We performed
a number of pre-processing steps on the reviews. These included removing
newlines and HTML tags, removing hyperlinks, replacing multiple dots
with one dot, and removing some special unicode characters such as
the heart symbol and special quotation symbols. Then any review containing
any character other than Arabic Unicode characters, numeric characters,
and punctuation is removed. Finally, any review that is composed of
only punctuation is also removed. This process filtered any review
containing non-Arabic characters and left us with 63,257 Arabic reviews.
The public release of the dataset includes only the cleaned up preprocessed
reviews in unicode format.

\subsection{Dataset Properties\label{sec:Dataset-Properties}}

The dataset contains 63,257 reviews that were submitted by 16,486
users for 2,131 different books. Table \ref{tab:Important-Dataset-Statistics.}
contains some important statistics about the dataset like the total
number of reviews in the dataset, the total number of users (reviewers),
the average reviews per user, median reviews per book, the total number
of books, average reviews per book, median tokens per review, maximum
tokens per review, average tokens per review, total number of tokens,
and total number of sentences.

Figure \ref{fig:Reviews-Histogram} shows the number of reviews for
each rating. The number of positive reviews is much larger than that
of negative reviews. We believe that this is because many of the reviewed
books are already popular books. The top rated books had many more
reviews, especially positive reviews, than the least popular books.
Figure \ref{fig:LABR-reviews-examples} shows some examples from the
data set, including long, medium, and short reviews. Notice the examples
colored in red, which represent problematic or noisy reviews. For
example, review 4 has positive sentiment text and negative rating,
while review 5 has negative sentiment text and positive rating. Notice
also the ambiguity for the reviews with rating 3, which can be associated
with positive, negative, or neutral. 

The average user provided 3.84 reviews with the median being 2. The
average book got 29.68 reviews with the median being 6. Figure \ref{fig:Number-of-reviews-per-user}
shows the number of reviews per user and book. By positive rating
we mean any review with rating more than 3 (4 and 5) and negative
rating means any review with rating lower than 3 (1 and 2). As shown
in the Figure \ref{fig:Number-of-reviews-per-user}\emph{c}, most
books and users have few reviews, and vice versa. Figures \ref{fig:Number-of-reviews-per-user}\emph{a-b}
show a box plot of the number of reviews per user and book for all,
positive, and negative reviews. We notice that books (and users) tend
to have (give) more positive reviews than negative reviews, where
the median number of positive reviews per book is 5 while that for
negative reviews is only 2. The median number of positive reviews
per user is 2 while that for negative reviews is only 1.

Figure \ref{fig:Tokens-and-Sentences} shows the statistics of tokens
and sentences. The reviews were tokenized using Qalsadi \footnote{available at \href{https://pypi.python.org/pypi/qalsadi}{pypi.python.org/pypi/qalsadi}}
and rough sentence counts were computed. The average number of tokens
per review is 33, the average number of sentences per review is 3.5,
and the average number of tokens per each sentence is 9. Figures \ref{fig:Tokens-and-Sentences}\emph{a-b}
show that the distribution is similar for positive and negative reviews.
Figure \ref{fig:Tokens-and-Sentences}\emph{c} shows a plot of the
frequency of the tokens in the vocabulary on a log-log scale, which
conforms to Zipf's law \citet{manning1999foundations}.

\section{Experiments}

In our previous work \citet{alyatyia}, we introduced the dataset
and performed a limited set of experiments for two tasks: binary sentiment
polarity classification and 5-way ratings classification. In this
work, we perform an extended survey of more classifiers typically
used for the sentiment polarity classification problem. In addition,
we add a new class to the sentiment polarity classification: the neutral
class. Moreover, we present a method for generating a sentiment lexicon
from the dataset and explore its effectiveness.

\subsection{Data Preparation\label{sub:Data-Preparation}}

\begin{figure}
\includegraphics[scale=0.2]{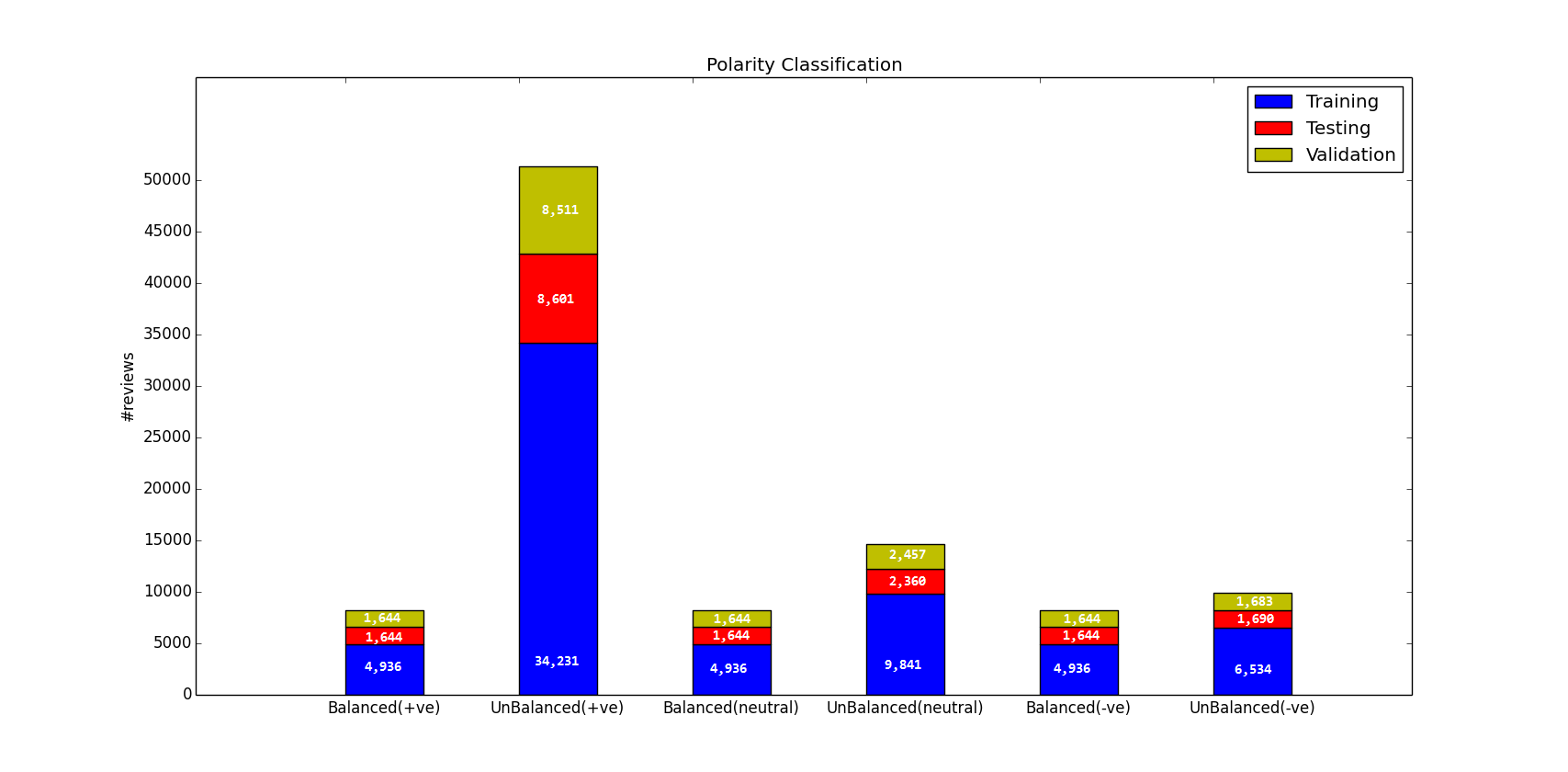}

\protect\caption{\textbf{Dataset Splits. }Number of reviews for each class category
for training, validation, and test sets for both balanced and unbalanced
settings.\label{fig:Number-of-reviews} See section \ref{sub:Data-Preparation}.}
\end{figure}

\begin{figure}
\includegraphics[scale=0.2]{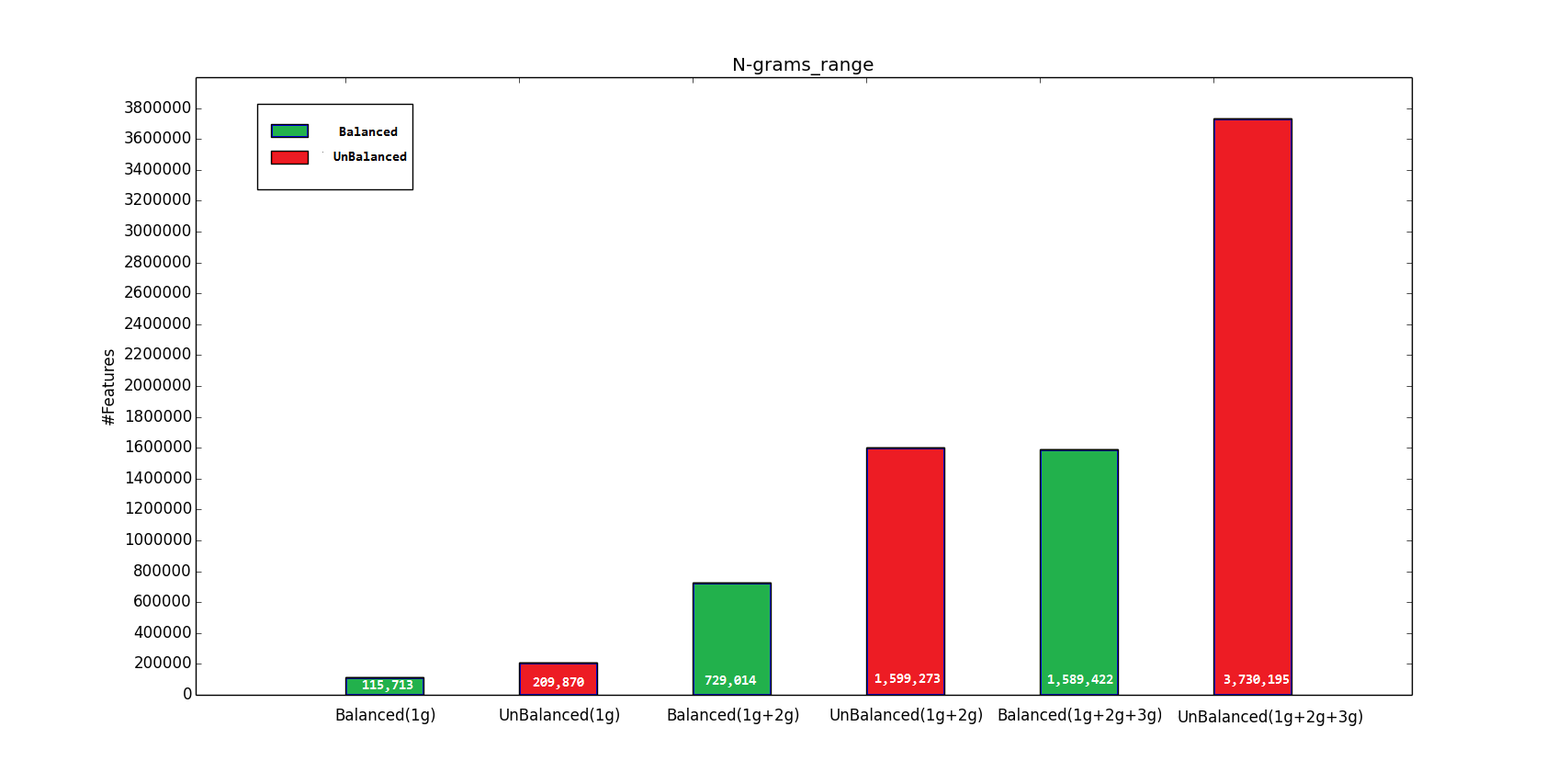}\protect\caption{\textbf{Feature Counts. }Number of unigram, bigram, and trigram features
per each class category.\label{fig:Number-of-features,} See section
\ref{sub:Data-Preparation}.}
\end{figure}

In order to test the proposed approaches thoroughly, we partition
the data into training, validation and test sets. The validation set
is used as a mini-test for evaluating and comparing models for possible
inclusion into the final model. The ratio of the data among these
three sets is 6:2:2 respectively.

We extend the work in \citep{alyatyia} by adding a class for neutral
reviews. In particular, instead of partitioning into just positive
and negative reviews, the data is divided into three classes (positive,
negative, and neutral) where ratings of 4 and 5 are mapped to positive,
rating of 3 is mapped to neutral, and ratings 1 and 2 are mapped to
negative. The neutral class is important, because some of the readers'
opinions are not swayed one way or the other towards positive or negative.
There is also some prevalence reviews that provide the positive and
the negative aspects, or simply provide an objective and neutral description.
We constructed two sets of data. The first one is the balanced data
set, where the number of reviews are equal in each class category,
by setting the size of the class to the minimum size of the three
classes. The second one is the unbalanced data set, where the number
of reviews are not equal, and their proportions match those of the
collected data set. Figure \ref{fig:Number-of-reviews} and Table
\ref{tab:Experiment-1-statistics} show the number of reviews for
each class category in the training, test, and validation sets for
both the balanced and unbalanced settings. Figure \ref{fig:Number-of-features,}
also shows the number of n-gram counts for both the balanced and unbalanced
settings. Notice the explosion in the size of features when using
unigrams, bigrams, and trigrams in the unbalanced setting, which exceeds
3.7 million features. This poses challenges in the training algorithms,
and provides a motivation for trying to reduce the feature dimension
using lexicons as explained in Section \ref{sec:Sentiment-Lexicon}.

\begin{table*}
\center
\scalebox{0.95}{%
\begin{tabular}{|c|c||c|c|c|c|c|c|}
\cline{3-8} 
\multicolumn{2}{c|}{\begin{turn}{90}
\end{turn}} & \multicolumn{3}{c|}{Balanced} & \multicolumn{3}{c|}{Unbalanced}\tabularnewline
\cline{3-8} 
\multicolumn{2}{c|}{\begin{turn}{90}
\end{turn}} & Positive & Negative & Neutral & Positive & Negative & Neutral\tabularnewline
\hline 
\multirow{3}{*}{Reviews Count} & Train Set & 4,936 & 4,936 & 4,936 & 34,231 & 6,534 & 9,841\tabularnewline
\cline{2-8} 
 & Test Set & 1,644 & 1,644 & 1,644 & 8,601 & 1,690 & 2,360\tabularnewline
\cline{2-8} 
 & Validation Set & 1,644 & 1,644 & 1,644 & 8,511 & 1,683 & 2,457\tabularnewline
\hline 
\multirow{3}{*}{Features Count} & unigrams & \multicolumn{3}{c|}{115,713} & \multicolumn{3}{c|}{209,870}\tabularnewline
\cline{2-8} 
 & unigrams+bigrams & \multicolumn{3}{c|}{729,014} & \multicolumn{3}{c|}{1,599,273}\tabularnewline
\cline{2-8} 
 & unigrams+bigrams+trigrams & \multicolumn{3}{c|}{1,589,422} & \multicolumn{3}{c|}{3,730,195}\tabularnewline
\hline 
\end{tabular}

}

\protect\caption{\textbf{Dataset Preparation Statistics}.\label{tab:Experiment-1-statistics}
The top part shows the number of reviews for the training, validation,
and test sets for each class category in both the balanced and unbalanced
settings. The bottom part shows the number of features. See Section
\ref{sub:Data-Preparation}.}
\end{table*}

\subsection{Sentiment Analysis \label{sub:Sentiment-Analysis}}

We explored using the dataset for two tasks: (a) \textbf{Sentiment
polarity classification: }where the goal is to predict if the review
is positive i.e. with rating 4 or 5, is negative i.e. with rating
1 or 2, or neutral with rating 3; and (b) \textbf{Rating classification}:
where the goal is to predict the rating of the review on a scale of
1 to 5.

In the two tasks a wide range of standard classifiers are applied
to both the balanced and unbalanced datasets using n-gram range of
all unigrams, bigrams and trigrams where the n-gram range of $N$
degree is a combination of all lower n-grams (contiguous sequence
of $n$ words) starting from unigrams, bigrams, ... etc. till the
degree $N$. For example the trigram range is a combination of unigrams,
bigrams and trigrams. Figure \ref{fig:Number-of-reviews} shows the
number of reviews in every class for both balanced and unbalanced
sets, while Figure \ref{fig:Number-of-features,} and Table \ref{tab:Experiment-1-statistics}
show the statistics of the number of features for uni-grams range,
bi-grams and trigrams range. The experiment is applied on both the
token counts and the Tf-Idf (token frequency inverse document frequency)
of the n-grams. Tf-Idf a way to normalize the document's word frequency
in a way that emphasizes words that are frequent or existing in the
current document, while being not frequent in the remaining documents
(see equation \ref{eq:tfidf}), and is defined as:

\begin{multline}
t(w,d)=\log(1+f(w,d))\times\log(\frac{D}{f(w)}),\\*
\label{eq:tfidf}
\end{multline}
 where $t(w,d)$ is the tf-idf weight for word $w$ in document $d$,
$f(w,d)$ is the frequency of word $w$ in document $d$, $D$ is
the total number of documents, and $f(w)$ is the total frequency
of word $w$.

The classifiers used in this experiment are widely used in the area
of sentiment analysis, and can be considered as a baseline benchmark
for any further experiments on the dataset. Python \textbf{scikit-learn}\footnote{http://scikit-learn.org/}
library is used for the experiments with default parameter settings
for each classifier. The classifiers are:
\begin{enumerate}
\item \textbf{Multinomial Naive Bayes(MNB): }A well-known method that is
used in many NLP tasks. In this method each review is represented
as a bag of words $\bar{X}=<x_{1},x_{2}...,x_{n}>$ where the feature
values are the term frequencies then the Bayes rule can be applied
to form a linear classifier.
\[
\log(p(class|\bar{X}))=\log(p(class)\times\prod_{i=1}^{n}\frac{p(x_{i}|class)}{p(\bar{X})})
\]

\item \textbf{Bernoulli Naive Bayes(BNB): }In this model\textbf{ }features
are independent binary variables that describe the input $\bar{X}=<1,0,1...1>$,
which means that the binary term occurrence is used instead of the
frequency of the term in the bag of words model. Both of the naive
Bayes generative models are described in details in \citet{mccallum1998comparison} 
\item \textbf{Support Vector Machine(SVM): }Linear SVM is a classifier that
partitions the data using the linear formula $y=\bar{W}\,.\,\bar{X}+p$,
selected in such a way that it maximizes the margin of separation
between the decision boundary and the class patterns (hence the name
large margin classifier). SVM can be generalized to multiclass case
using one versus all classification trick.
\item \textbf{Passive Aggressive:} It is an online learning model that uses
a hinge-loss function together with an aggressiveness parameter $C$,
in order to achieve a positive margin high confidence classifier.
The algorithm is described in details in \citep{crammer2006online}\textbf{
}with two alternative modifications that improve the algorithm's ability
to cope with noise.
\item \textbf{Stochastic Gradient Descent(SGD): }It is an algorithm that
is used to train other machine learning algorithms such as SVM where
it samples a subset of the training examples at every learning step.
Then it calculates the gradient from this subset only, and uses this
gradient to update the weight vector $w$ of SVM classifier. Because
of its simplicity and computational advantage, it is widely used for
large-scale machine learning problems\citep{bottou2007tradeoffs}.
\item \textbf{Logistic Regression: }The binary logistic regression uses
a sigmoid function $h_{w}(x)=f(x)=1/(1+e^{-w^{T}x})$ as a learning
model, then it optimizes a cost function that measures the likelihood
of the data given the classifier's class probability estimates then
for the multiclass problem one versus all solution is used. The cost
function can be formulated as
\begin{eqnarray*}
Cost(\bar{w}) & = & \frac{-1}{m}\sum_{i=1}^{n}[y^{(i)}log(h_{w}(x^{(i)}))+\\
 &  & (1-y^{(i)})log(1-h_{w}(x^{(i)}))]
\end{eqnarray*}
 where $m$ is the total number of patterns, $x^{(i)}$is the $i^{th}$
pattern and $y^{(i)}$ is the correct class of the pattern $i$. 
\item \textbf{Linear Perceptron: }It is a simple feed-forward single layer
linear neural network with a unit step function as an activation function.
It uses an iterative algorithm for training the weights. However,
this algorithm does not take into account the margin like for the
case of SVM.
\item \textbf{K-Nearest Neighbor(KNN):} A simple well-known machine learning
classifier that based on the distances between the patterns in the
feature space. Specifically, a pattern is classified according to
the majority class of its K-nearest neighbors. 
\end{enumerate}
\begin{table*}
\center
\scalebox{1.1}{

\begin{tabular}{|c|c|c|c|c||c|c|c|}
\cline{3-8} 
\multicolumn{1}{c}{} &  & \multicolumn{3}{c||}{Precision} & \multicolumn{3}{c|}{Recall}\tabularnewline
\cline{3-8} 
\multicolumn{1}{c}{} &  & Positive & Negative & Neutral & Positive & Negative & Neutral\tabularnewline
\hline 
\multirow{2}{*}{SVM} & Balanced & 0.62639 & 0.64216 & 0.4963 & 0.64936 & 0.64841 & 0.46836\tabularnewline
\cline{2-8} 
 & UnBalanced & \textbf{0.79476} & 0.68744 & 0.48185 & \textbf{0.93291} & 0.46982 & 0.28686\tabularnewline
\hline 
\end{tabular}

}

\protect\caption{\textbf{SVM Classifier Results} .\label{tab:SVM-Classifier-Results}
The table shows the precision and the recall of the SVM classifier
when evaluate on both balanced and unbalanced test sets and trained
using TF-IDF trigram range of features.}
\end{table*}

Table \ref{tab:Experiment-2:-Polarity}-\ref{tab:Rating-Classification-Results}
shows the result for each classifier after training on both the training
and the validation set and evaluating the result on the test set (i.e.
the train:test ratio is 8:2). Each cell has numbers that represent
\textbf{weighted}\textbf{\emph{ }}\textbf{accuracy / F1 measure} where
the evaluation is performed on the test set. Table \ref{tab:Experiment-2:-Polarity}
shows the results of the polarity classification task while Table\ref{tab:Rating-Classification-Results}
shows the results of the rating classification task. Note that in
the sentiment polarity classification task the inclusion of a third
class \textquotedbl{}neutral\textquotedbl{} makes the problem much
harder, and we get a lower performance than the case of two-class
case (\textquotedbl{}positive\textquotedbl{} and \textquotedbl{}negative\textquotedbl{}).
The reason is that there is a large confusion between the neutral
class and both the positive and negative classes. Sometimes the numbered
ratings (1 to 5), from which we extract the target class labels, contradict
what is written in the review, in a way that even an experienced human
analyzer will not get it right (Examples are marked in red in Fig
\ref{fig:LABR-reviews-examples}). Two accuracy measures are used
to calculate the performance. The first is the weighted accuracy \ref{eq:acc} 

\begin{equation}
a=\sum_{c=1}^{C}a(c)w(c)\label{eq:acc}
\end{equation}
 where $a$ is the weighted accuracy, $C$ is the number of classes,
\[
w(c)=\frac{n(c)}{\sum_{c}n(c)}
\]
 is the weight for class $c$,$n(c)$ is the number of reviews in
class $c$, and $a(c)$ is the accuracy of class $c$ defined as:

\[
a(c)=\frac{tp(c)}{n(c)}
\]
where $tp(c)$ is the number of true positives for class $c$ (the
number of reviews that are algorithm correctly identified as class
$c$). The second measure is the weighted F1 measure \ref{eq:fq}:

\begin{equation}
F1=\sum_{c}\frac{2p(c)\times r(c)}{p(c)+r(c)}w(c)\label{eq:fq}
\end{equation}
where $w(c)$ is the weight for class $c$ as defined above, $p(c)$
is the precision for class $c$ and $r(c)$ is its recall defined
as 

\[
p(c)=\frac{tp(c)}{tp(c)+fp(c)}
\]

\[
r(c)=\frac{tp(c)}{tp(c)+fn(c)}
\]
with $fp(c)$ the false positives and $fn(c)$ the false negatives
for class $c$.

\begin{table*}
\center
\scalebox{0.95}{

\begin{tabular}{|c|c||c|c|c||c|c|c|}
\hline 
\multirow{2}{*}{{\footnotesize{}Features}} & \multirow{2}{*}{{\footnotesize{}Tf-Idf}} & \multicolumn{3}{c||}{{\footnotesize{}Balanced}} & \multicolumn{3}{c|}{{\footnotesize{}Unbalanced}}\tabularnewline
\cline{3-8} 
 &  & {\footnotesize{}1g} & {\footnotesize{}1g+2g} & {\footnotesize{}1g+2g+3g} & {\footnotesize{}1g} & {\footnotesize{}1g+2g} & {\footnotesize{}1g+2g+3g}\tabularnewline
\hline 
\hline 
\multirow{2}{*}{{\footnotesize{}MNB}} & {\footnotesize{}No} & 0.558/0.560 & 0.573/0.577 & 0.572/0.577 & 0.706/0.631 & 0.705/0.609 & 0.706/0.612\tabularnewline
\cline{2-8} 
 & {\footnotesize{}Yes} & 0.567/0.570 & 0.581/0.584 & 0.582/0.586 & 0.680/0.551 & 0.680/0.550 & 0.680/0.550\tabularnewline
\hline 
\multirow{2}{*}{{\footnotesize{}BNB}} & {\footnotesize{}No} & 0.515/0.495 & 0.507/0.473 & 0.481/0.429 & 0.659/0.573 & 0.674/0.553 & 0.678/0.550\tabularnewline
\cline{2-8} 
 & {\footnotesize{}Yes} & 0.356/0.236 & 0.341/0.189 & 0.338/0.181 & \multicolumn{1}{c|}{0.680/0.550} & 0.680/0.550 & 0.680/0.550\tabularnewline
\hline 
\multirow{2}{*}{{\footnotesize{}SVM}} & {\footnotesize{}No} & 0.535/0.534 & 0.568/0.565 & 0.570/0.566 & 0.698/0.690 & 0.727/0.712 & 0.731/0.712\tabularnewline
\cline{2-8} 
 & {\footnotesize{}Yes} & 0.566/0.564 & \textbf{0.590/0.588} & \textbf{0.589/0.588} & \textbf{0.734/0.709} & \textbf{0.750/0.723} & \textbf{0.751/0.725}\tabularnewline
\hline 
\multirow{2}{*}{Passive Aggressive} & {\footnotesize{}No} & 0.402/0.348 & 0.489/0.486 & 0.521/0.525 & 0.638/0.653 & 0.693/0.692 & 0.692/0.676\tabularnewline
\cline{2-8} 
 & {\footnotesize{}Yes} & 0.504/0.508 & 0.571/0.574 & 0.584/0.582 & 0.681/0.676 & 0.740/0.722 & 0.740/0.715\tabularnewline
\hline 
\multirow{2}{*}{SGD} & {\footnotesize{}No} & 0.458/0.454 & 0.459/0.454 & 0.459/0.455 & 0.687/0.578 & 0.687/0.579 & 0.680/0.570\tabularnewline
\cline{2-8} 
 & {\footnotesize{}Yes} & 0.416/0.390 & 0.380/0.292 & 0.360/0.236 & 0.680/0.550 & 0.680/0.550 & 0.673/0.541\tabularnewline
\hline 
\multirow{2}{*}{Logistic Regression} & {\footnotesize{}No} & 0.570/0.568 & 0.586/0.583 & 0.590/0.585 & 0.728/0.707 & 0.743/0.717 & 0.737/0.703\tabularnewline
\cline{2-8} 
 & {\footnotesize{}Yes} & \textbf{0.587/0.583} & \textbf{0.590/0.588} & 0.586/0.585 & 0.727/0.672 & 0.720/0.659 & 0.709/0.640\tabularnewline
\hline 
\multirow{2}{*}{Linear Perceptron} & {\footnotesize{}No} & 0.389/0.328 & 0.424/0.375 & 0.449/0.418 & 0.683/0.680 & 0.720/0.705 & 0.719/0.693\tabularnewline
\cline{2-8} 
 & {\footnotesize{}Yes} & 0.500/0.502 & 0.536/0.538 & 0.526/0.523 & 0.675/0.672 & 0.732/0.714 & 0.726/0.708\tabularnewline
\hline 
\multirow{2}{*}{KNN} & {\footnotesize{}No} & 0.428/0.416 & 0.412/0.395 & 0.398/0.382 & 0.675/0.582 & 0.676/0.577 & 0.673/0.567\tabularnewline
\cline{2-8} 
 & {\footnotesize{}Yes} & 0.471/0.461 & 0.497/0.484 & 0.490/0.477 & 0.698/0.619 & 0.701/0.625 & 0.697/0.615\tabularnewline
\hline 
\end{tabular}

}\protect\caption{\textbf{Experiment 1: Polarity Classification Experimental Results}\label{tab:Polarity-Classification-Results}.\emph{Tf-Idf}
indicates whether tf-idf weighting was used or not. \emph{MNB} is
Multinomial Naive Bayes, \emph{BNB} is Bernoulli Naive Bayes,\emph{
SVM} is the Support Vector Machine, SGD is the stochastic gradient
descent and KNN is the K-nearest neighbor. The numbers represent weighted\emph{
}accuracy / F1 measure where the evaluation is performed on the test
set.\label{tab:Experiment-2:-Polarity} For example, 0.558/0.560 means
a weighted accuracy of 0.558 and an F1 score of 0.560.}
\end{table*}

\begin{table*}
\center
\scalebox{0.95}{

\begin{tabular}{|c|c||c|c|c||c|c|c|}
\hline 
\multirow{2}{*}{{\footnotesize{}Features}} & \multirow{2}{*}{{\footnotesize{}Tf-Idf}} & \multicolumn{3}{c||}{{\footnotesize{}Balanced}} & \multicolumn{3}{c|}{{\footnotesize{}Unbalanced}}\tabularnewline
\cline{3-8} 
 &  & {\footnotesize{}1g} & {\footnotesize{}1g+2g} & {\footnotesize{}1g+2g+3g} & {\footnotesize{}1g} & {\footnotesize{}1g+2g} & {\footnotesize{}1g+2g+3g}\tabularnewline
\hline 
\hline 
\multirow{2}{*}{{\footnotesize{}MNB}} & {\footnotesize{}No} & 0.390/0.394 & 0.408/0.416 & 0.409/0.416 & 0.459/0.421 & 0.470/0.416 & 0.474/0.418\tabularnewline
\cline{2-8} 
 & {\footnotesize{}Yes} & 0.399/0.403 & 0.420/0.299 & 0.420/0.299 & 0.416/0.301 & 0.427/0.430 & 0.428/0.431\tabularnewline
\hline 
\multirow{2}{*}{{\footnotesize{}BNB}} & {\footnotesize{}No} & 0.330/0.296 & 0.304/0.254 & 0.269/0.202 & 0.408/0.331 & 0.393/0.263 & 0.386/0.236\tabularnewline
\cline{2-8} 
 & {\footnotesize{}Yes} & 0.223/0.125 & 0.222/0.184 & 0.205/0.279 & \multicolumn{1}{c|}{0.376/0.206} & 0.376/0.206 & 0.376/0.206\tabularnewline
\hline 
\multirow{2}{*}{{\footnotesize{}SVM}} & {\footnotesize{}No} & 0.377/0.374 & 0.396/0.388 & 0.400/0.392 & 0.467/0.461 & 0.489/0.480 & 0.495/0.483\tabularnewline
\cline{2-8} 
 & {\footnotesize{}Yes} & 0.395/0.392 & 0.417/0.412 & 0.420/0.414 & \textbf{0.487/0.477} & \textbf{0.513/0.500} & \textbf{0.519/0.505}\tabularnewline
\hline 
\multirow{2}{*}{Passive Aggressive} & {\footnotesize{}No} & 0.279/0.233 & 0.339/0.305 & 0.363/0.338 & 0.427/0.429 & 0.471/0.459 & 0.460/0.454\tabularnewline
\cline{2-8} 
 & {\footnotesize{}Yes} & 0.360/0.359 & 0.398/0.388 & 0.399/0.388 & 0.449/0.447 & 0.499/0.486 & 0.511/0.494\tabularnewline
\hline 
\multirow{2}{*}{SGD} & {\footnotesize{}No} & 0.210/0.194 & 0.210/0.193 & 0.212/0.198 & 0.439/0.431 & 0.483/0.471 & 0.484/0.477\tabularnewline
\cline{2-8} 
 & {\footnotesize{}Yes} & 0.202/0.171 & 0.200/0.167 & 0.200/0.167 & 0.482/0.440 & 0.502/0.466 & 0.509/0.477\tabularnewline
\hline 
\multirow{2}{*}{Logistic Regression} & {\footnotesize{}No} & 0.391/0.386 & 0.414/0.405 & 0.420/0.410 & 0.487/0.475 & 0.506/0.492 & 0.512/0.495\tabularnewline
\cline{2-8} 
 & {\footnotesize{}Yes} & \textbf{0.410/0.404} & \textbf{0.429/0.424} & \textbf{0.433/0.430} & 0.484/0.455 & 0.497/0.461 & 0.495/0.457\tabularnewline
\hline 
\multirow{2}{*}{Linear Perceptron} & {\footnotesize{}No} & 0.242/0.179 & 0.271/0.220 & 0.304/0.265 & 0.448/0.445 & 0.478/0.473 & 0.490/0.475\tabularnewline
\cline{2-8} 
 & {\footnotesize{}Yes} & 0.358/0.351 & 0.397/0.384 & 0.398/0.384 & 0.441/0.440 & 0.490/0.476 & 0.492/0.479\tabularnewline
\hline 
\multirow{2}{*}{KNN} & {\footnotesize{}No} & 0.256/0.233 & 0.259/0.242 & 0.257/0.240 & 0.342/0.336 & 0.341/0.333 & 0.339/0.334\tabularnewline
\cline{2-8} 
 & {\footnotesize{}Yes} & 0.298/0.284 & 0.308/0.295 & 0.316/0.305 & 0.374/0.368 & 0.386/0.375 & 0.392/0.381\tabularnewline
\hline 
\end{tabular}

}\protect\caption{\textbf{Rating Classification Experimental Results} The numbers represent
weighted\emph{ }accuracy / F1 measure where the evaluation is performed
on the test set.\label{tab:Rating-Classification-Results}}
\end{table*}

From tables \ref{tab:Experiment-2:-Polarity} and \ref{tab:Rating-Classification-Results}
we can make the following observations:
\begin{enumerate}
\item The ratings classification task is more challenging than the polarity
classification task. This is to be expected, since we are dealing
with five classes in the former, as opposed to only three in the latter.
\item The balanced set is more challenging than the unbalanced set for both
tasks. We believe that this is due to the fact that it contains much
fewer reviews compared to the unbalanced set. This makes a lot of
the ngrams have fewer training examples, and therefore leads to less
reliable classification. Table \ref{tab:SVM-Classifier-Results} shows
the precision and the recall of the SVM classifier when evaluate on
both balanced and unbalanced test sets and trained using TF-IDF trigram
range of features. Despite the fact that the overall performance of
the unbalanced dataset is better than the balanced dataset, the individual
performance of each class category in the unbalanced evaluation is
proportional to its ratio in the dataset.
\item We can get a good overall accuracy and good F1 using especially the
SVM and the logistic regression classifiers (over 70\% for the polarity
classification task in Table \ref{tab:Experiment-2:-Polarity}). This
is consistent with previous results in \citet{alyatyia} suggesting
that the SVM and the logistic regression are a reliable choice.
\item Passive aggressive, and linear perceptron are also a good choice of
classifiers, with the careful choice of parameters.
\end{enumerate}

\section{Sentiment Lexicon\label{sec:Sentiment-Lexicon}}

\subsection{Seed Lexicon Generation }

\begin{table*}
\center\includegraphics[width=0.8\paperwidth]{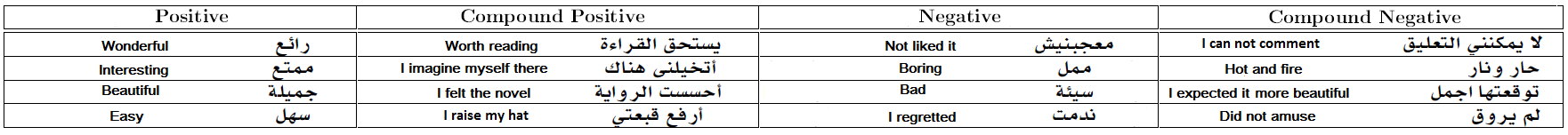}

\protect\caption{\textbf{Sentiment Lexicon Examples}\label{tab:Lexicon}. Notice how
our lexicon is able to automatically capture some difficult compound
terms from the training set. Notice also that some words that are
compound in English are actually one word in Arabic. See Section \ref{sec:Sentiment-Lexicon}.}
\end{table*}

\begin{table*}
\center
\scalebox{0.95}{

\begin{tabular}{|c|c||c|c|c|c|c|c|c|}
\hline 
\multirow{1}{*}{{\footnotesize{}Features}} & \multirow{1}{*}{{\footnotesize{}Tf-Idf}} & \multicolumn{1}{c|}{LEX1} & LEX2 & LEX1+ Trigrams & LEX2+ Trigrams & LEX1+ LEX2 & LEX1+ LEX2+Trigrams & Trigrams\tabularnewline
\hline 
\multirow{1}{*}{{\footnotesize{}MNB}} & {\footnotesize{}Yes} & 0.705/0.623 & 0.684/0.570 & 0.681/0.552 & 0.680/0.551 & \textbf{0.707/0.638} & 0.681/0.553 & 0.680/0.550\tabularnewline
\hline 
\multirow{1}{*}{{\footnotesize{}BNB}} & {\footnotesize{}Yes} & 0.696/0.627 & 0.675/0.576 & 0.680/0.550 & 0.680/0.550 & 0.690/0.627 & 0.680/0.550 & 0.680/0.550\tabularnewline
\hline 
\multirow{1}{*}{{\footnotesize{}SVM}} & {\footnotesize{}Yes} & \textbf{0.705/0.634} & 0.681/0.597 & \textbf{0.752/0.727} & 0.747/0.724 & 0.704/0.646 & 0.748/0.724 & 0.751/0.725\tabularnewline
\hline 
\multirow{1}{*}{Passive Aggressive} & {\footnotesize{}Yes} & 0.655/0.617 & 0.646/0.597 & \textbf{0.741}/0.723 & 0.739/\textbf{0.724} & 0.643/0.621 & 0.735/0.720 & 0.740/0.715\tabularnewline
\hline 
\multirow{1}{*}{SGD} & {\footnotesize{}Yes} & 0.699/0.608 & 0.685/0.580 & 0.715/0.635 & 0.695/0.601 & 0.705/0.625 & \textbf{0.719/0.649} & 0.673/0.541\tabularnewline
\hline 
\multirow{1}{*}{Logistic Regression} & {\footnotesize{}Yes} & 0.704/0.630 & \textbf{0.688/0.593} & \textbf{0.731/0.684} & 0.718/0.665 & 0.710/0.646 & 0.728/0.684 & 0.709/0.640\tabularnewline
\hline 
\multirow{1}{*}{Linear Perceptron} & {\footnotesize{}Yes} & 0.597/0.589 & 0.400/0.437 & \textbf{0.732}/0.716 & 0.730/\textbf{0.718} & 0.511/0.539 & 0.721/0.711 & 0.726/0.708\tabularnewline
\hline 
\multirow{1}{*}{KNN} & {\footnotesize{}Yes} & 0.642/0.610 & 0.622/0.576 & 0.654/\textbf{0.636} & 0.639/0.618 & 0.543/0.552 & 0.650/0.632 & \textbf{0.697/}0.615\tabularnewline
\hline 
\end{tabular}

}\protect\caption{\textbf{Sentiment lexicon experimental results}.\textbf{ }The numbers
represent weighted\emph{ }accuracy / F1 measure where the evaluation
is on the test set for the sentiment polarity classification (compare
with Table \ref{tab:Polarity-Classification-Results}). LEX1 indicates
our generated lexicon, LEX2 indicates the lexicon by \citet{el2013open},
and Trigrams indicates the trigram range features from the training
set. \textbf{\label{tab:Sentiment-lexicon-Experimental} }See Section
\ref{sec:Sentiment-Lexicon}.}
\end{table*}

Manually constructing a sentiment lexicon is a formidable task due
to the coverage issues and the possible ambiguity and multiple meanings
of many words. Also compound phrases open up many permutations of
word combinations which will be hard to group in the lexicon. So we
propose a simple method for extracting a seed sentiment lexicon from
the LABR dataset. This lexicon can be extended easily to other datasets
or domains. Our method utilizes a useful feature of the linear SVM
and logistic regression as they inherently apply some sort of feature
selection. This is because the weight values are an indication of
the importance of the n-gram. For example, n-grams that have negligible
weights are deemed unimportant and ineffective. This is especially
true if we use the $\ell_{1}$ error measure for training the SVM
(defined as $||x||_{1}=\sum_{i}|x_{i}|$). In this case, the weights
for many insignificant ngrams will end up being zero. So, we utilize
this fact to perform an automatic generation for the most informative
ngrams by ordering the weights from the SVM and the logistic regression
classifiers then selecting the highest 1000 weights, as indication
for positive sentiment ngrams, and the lowest 1000 weights as indication
for negative sentiment ngrams. We then manually review them to remove
any erroneous n-grams. We end up with a list of 348 negative n-grams
and 319 positive n-grams. We also constructed a list of 31 Arabic
negation operators. This lexicon can be considered a seed, and is
a first step in contructing a complete sentiment lexicon. As mentioned
\citet{elsahar2015building}, have independently proposed an idea
with some similarities (i.e. using SVM with L1 error measure). This
study has come out very recently. Table \ref{tab:Lexicon} gives some
examples from the sentiment lexicon where it is clear that some difficult
compound phrases were captured using our approach.

\subsection{Lexicon Experiments}

In order to test the effectiveness of the generated domain specific
lexicon, we reran the sentiment polarity classification experiments
on the unbalanced training set. The goal is to test the effectiveness
of the lexicon as a stand-alone input, and also in combination with
the trigram features used in the previous experiments. The lexicon
was used as a feature vector of length 667 features (348 negative
ngrams and 319 positive ngrams). If using the lexicon as stand-alone
is as successful, then we would have reduced the number of features
from several millions to just 667, leading to a much simpler classifier.
Moreover, this opens up the possibility of using some complex classifiers
that were computationally unfeasible with large feature vectors. These
classifiers may perhaps outperform the simpler classifiers typically
used in large NLP problems. As a comparison, we consider the Arabic
sentiment lexicon developed by \citet{el2013open} (see also \citet{ElSaharE14}
for more details of its construction). This is a general purpose lexicon
that is developed by growing a small seed of manually labeled words
using an algorithm that considers co-occurrence of words in the text.
It consists of 4392 entries of both compound and single sentiment
words. Table \ref{tab:Sentiment-lexicon-Experimental} shows the results
on the test set, where we use a combined training and validation set
for training the models. We can observe that the lexicon only model
is just a little worse than the trigram model. However, the difference
is not large. This is an interesting fact considering that the former
uses only 0.02 \% of the amount of features of the latter. Another
observation is that our constructed lexicon outperforms the lexicon
by \citet{el2013open}. But this has mainly to do with the fact that
ours is domain-specific, while theirs is general purpose. There are
many entries in our lexicon that are specific to the book review domain,
and can therefore make a difference in performance. For example, see
in Table \ref{tab:Lexicon} the expressions \textquotedbl{}worth reading\textquotedbl{},
\textquotedbl{}I imagine myself there\textquotedbl{}, and \textquotedbl{}I
felt the novel\textquotedbl{}. This indicates that it is always a
good practice to augment general purpose lexicons with domain-specific
expressions. Notice that some terms that are compound in English are
actually represented by one word in Arabic, for example the first
and last rows in the negatives in Table \ref{tab:Lexicon}.

\section{Summary and Conclusion}

In this work we presented LABR dataset the largest Arabic sentiment
analysis dataset to-date. We explored its properties and statistics,
and provided standard splits. We performed a comprehensive study,
involving testing a wide range of classifiers to provide a baseline
for future comparisons. We also presented a small sentiment lexicon
that is extracted from the dataset and explored its effectiveness.
From the experiments we observe that SVM and logistic regression are
the best two classifiers. Moreover, using the constructed lexicon
we obtained competitive results with only a fraction of the number
of features. We hope this data set would be of good use, and these
results would be a guide for future Arabic sentiment work.

\section*{Acknowledgements }

Some of this work has been funded by ITIDA's ITAC project number CFP-65.

\bibliographystyle{abbrvnat}
\bibliography{Sentiment}

\end{document}